\pdfoutput=1

\documentclass[11pt]{article}

\usepackage{acl}

\usepackage{times}
\usepackage{latexsym}
\usepackage{xcolor}
\usepackage{booktabs}

\usepackage{algorithm}
\usepackage{algorithmic}
\makeatletter
\newcommand{\algorithmictry}{\textbf{try}}
\newcommand{\algorithmiccatch}{\textbf{catch}}
\newcommand{\algorithmicendtry}{\textbf{end try}}
\newcommand{\TRY}{\ALC@it\algorithmictry\begin{ALC@if}}
\newcommand{\CATCH}{\end{ALC@if}\ALC@it\algorithmiccatch\begin{ALC@if}}
\newcommand{\ENDTRY}{\end{ALC@if}\ALC@it\algorithmicendtry}
\makeatother
\usepackage{amsmath}
\usepackage{amssymb}
\usepackage{tcolorbox}
\tcbuselibrary{breakable}
\usepackage{float}
\newtcolorbox{promptbox}{
  breakable,
  colback=gray!5!white,
  colframe=gray!75!black,
  title=LLM Prompt
}

\graphicspath{{./}}
\usepackage{mdframed}
\usepackage{placeins}
\usepackage{hyperref}
\hypersetup{
  colorlinks=true,
  linkcolor=black,
  citecolor=blue,
  urlcolor=blue,
  pdftitle={Rushes: A Human Preference Dataset for Pluralistic Alignment},
  pdfauthor={Michael Xu, Jorge Leandro, Sudha Rao, Weijia Xu, Nebojsa Jojic, Gabriel DesGarennes, Chris Quirk, Bill Dolan},
  pdfsubject={A dataset and benchmark for revealed engagement preferences in interactive narratives},
  pdfkeywords={pluralistic alignment, human preferences, interactive narratives, personalization}
}
\usepackage{listings}
\usepackage{tikz}
\usetikzlibrary{arrows.meta, positioning}
\usepackage[T1]{fontenc}

\usepackage[utf8]{inputenc}

\usepackage{microtype}

\usepackage{inconsolata}

\usepackage{graphicx}

\usepackage{cuted}

\newcommand{\adjustimg}{%
  \hspace*{\dimexpr\evensidemargin-\oddsidemargin}%
}
\newcommand{\centerimg}[2][width=\textwidth]{%
  \makebox[\textwidth]{\adjustimg\includegraphics[#1]{#2}}%
}

\title{\textit{Rushes}: A Human Preference Dataset for Pluralistic Alignment}

\author{Michael Xu \quad Jorge Leandro \quad Sudha Rao \quad Weijia Xu \\ \textbf{Nebojsa Jojic} \quad \textbf{Gabriel DesGarennes} \quad \textbf{Chris Quirk} \quad  \textbf{Bill Dolan} \\ Microsoft Research \\
  \texttt{michaelxu@microsoft.com} }

\begin{document}
\maketitle

\begin{strip}
  \vspace{-58pt}

    \noindent\centerimg[width=\linewidth]{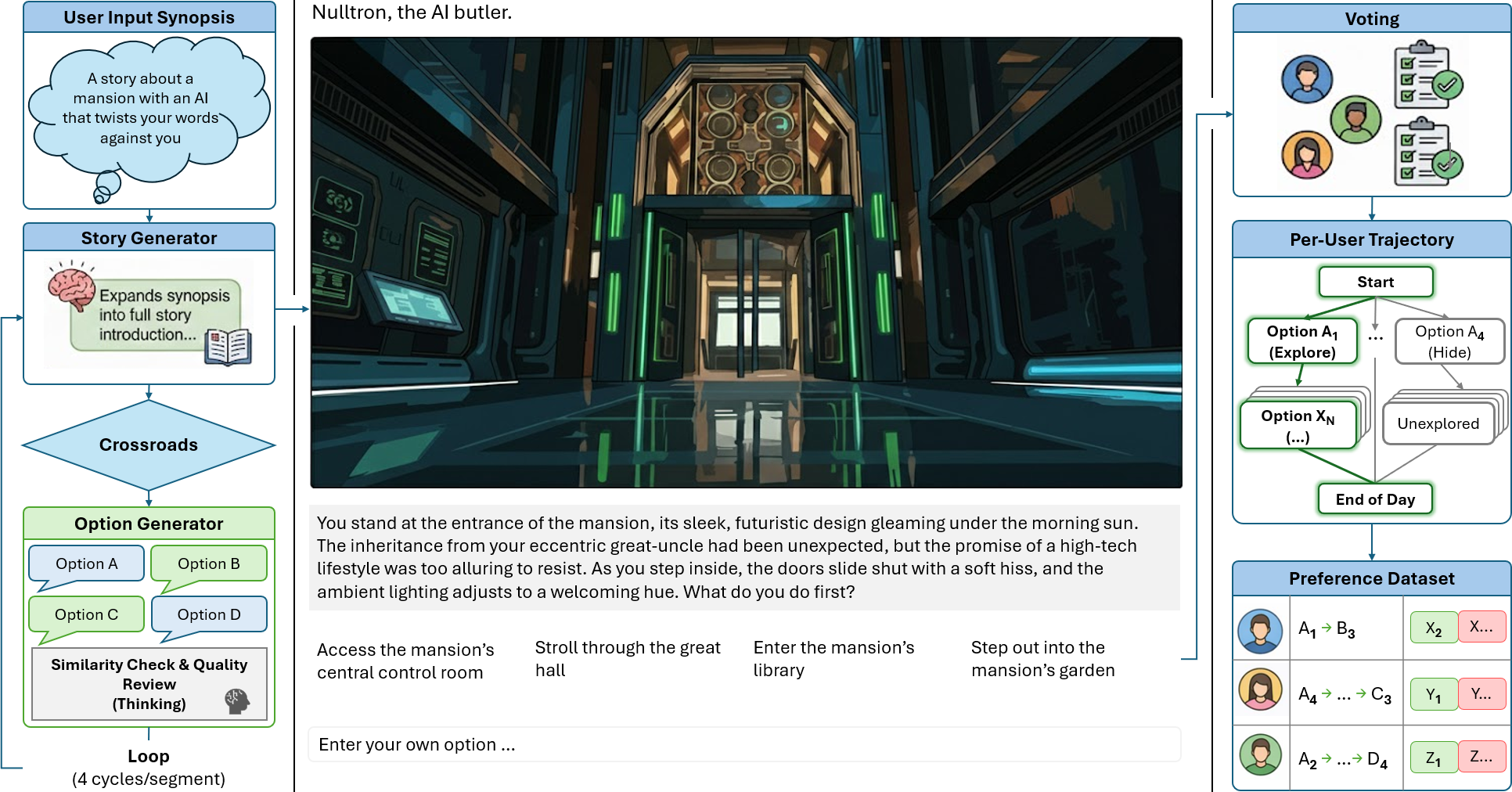}

  \refstepcounter{figure}
  \label{fig:teaser}
  \fontsize{10pt}{12pt}\selectfont
    Figure \thefigure: Overview of the \textit{Rushes} data collection framework. \textbf{Left:} The generation pipeline expands a user-provided synopsis into story segments and iteratively generates distinct options using similarity and quality checks. \textbf{Center:} The game interface presents the narrative and visual context, allowing players to select predefined choices or submit their own suggestions. \textbf{Right:} The data collection mechanism aggregates user votes and individual playthrough trajectories into a revealed-preference interaction dataset.

\end{strip}

\begin{abstract}
    We introduce \textbf{Rushes}, a dataset and benchmark for studying revealed human engagement preferences in interactive narrative environments. \textit{Rushes} was collected through a game interface in which users interacted with AI-generated branching narratives and selected one choice from a small, explicit candidate set at each decision point. Each interaction logs the full candidate set, the user's choice, and the evolving narrative context, yielding time-ordered trajectories with persistent user-level identifiers.

    \textit{Rushes} contains \textbf{44,226 decision events from 8,167 unique users across six games}, capturing sequential, personalized engagement behavior rather than static judgments. We show that user choices exhibit structured, non-random patterns, quantified by lower mean choice entropy than a uniform four-choice baseline.

    We position \textit{Rushes} as a diagnostic benchmark for pluralistic alignment and demonstrate an \textit{Engagement Gap}: state-of-the-art LLMs, including GPT-5, do not outperform simple baselines. An SVD-based matrix factorization model captures measurable personalized signal (37.7\%), whereas GPT-5 with user history reaches 34.2\%, below the popularity baseline at 36.4\%, on event-level choice prediction. This gap suggests that population-level objectives, such as those used in modern RLHF, may be insufficient to capture heterogeneous, context-dependent engagement signals. Even highly capable models may therefore default to majority preferences rather than adapt to individual trajectories. We release \textit{Rushes} to support research into pluralistic alignment and sequential decision-making in generative systems.
\end{abstract}

\section{Introduction}

Foundational work on large language models (LLMs) remains largely focused on capability and safety, codified by datasets that reward helpful and harmless outputs. In both safety research and practice, the goal is often convergence: to minimize harm for all users.
Entertainment domains such as games, movies, or books, however, introduce an orthogonal dimension to model development. Here the goal is divergence: to maximize ``interestingness'' and ``fun'' for specific individuals. Learning what makes an experience meaningful across subjective dimensions will be critical in applications with multiple valid targets for different users.

A personalized notion of engagement calls for pluralistic alignment, in which models adapt to diverse human values rather than collapse to a single mean. Current alignment methods, however, often fail to capture this subjectivity. As noted by \citet{ali2025operationalizingpluralisticvalueslarge}, aggregating diverse preferences into a single reward model suppresses minority viewpoints, leading to generic outputs. Furthermore, contextual history plays an important role in these settings because prior user choices can inform subsequent model predictions.

To our knowledge, no prior large-scale human preference dataset jointly addresses engagement alignment, sequential decision-making, and personalized modeling. We present \textit{Rushes}, a dataset and benchmark built around human responses to AI-generated branching narratives that include text, images, video, and audio narration.

\textbf{The Engagement Gap:} Our experiments reveal a critical limitation in current frontier models. When tasked with predicting user choices in \textit{Rushes}, models such as GPT-4o \cite{openai2024gpt4o} and GPT-5 \cite{openai2025gpt5} do not outperform simple popularity heuristics. This mirrors popularity bias in recommender systems but highlights a distinct failure mode in LLMs: they are fine-tuned to be ``universally acceptable'' rather than ``personally compelling.'' Recent work by \citet{castricato-etal-2025-persona} with the \textit{PERSONA} benchmark has begun to address this problem using synthetic user proxies. \textit{Rushes} complements this synthetic approach with organic, revealed preferences from human trajectories, capturing the noisy and implicit nature of engagement.

\begin{figure}[t]
    \includegraphics[width=\columnwidth]{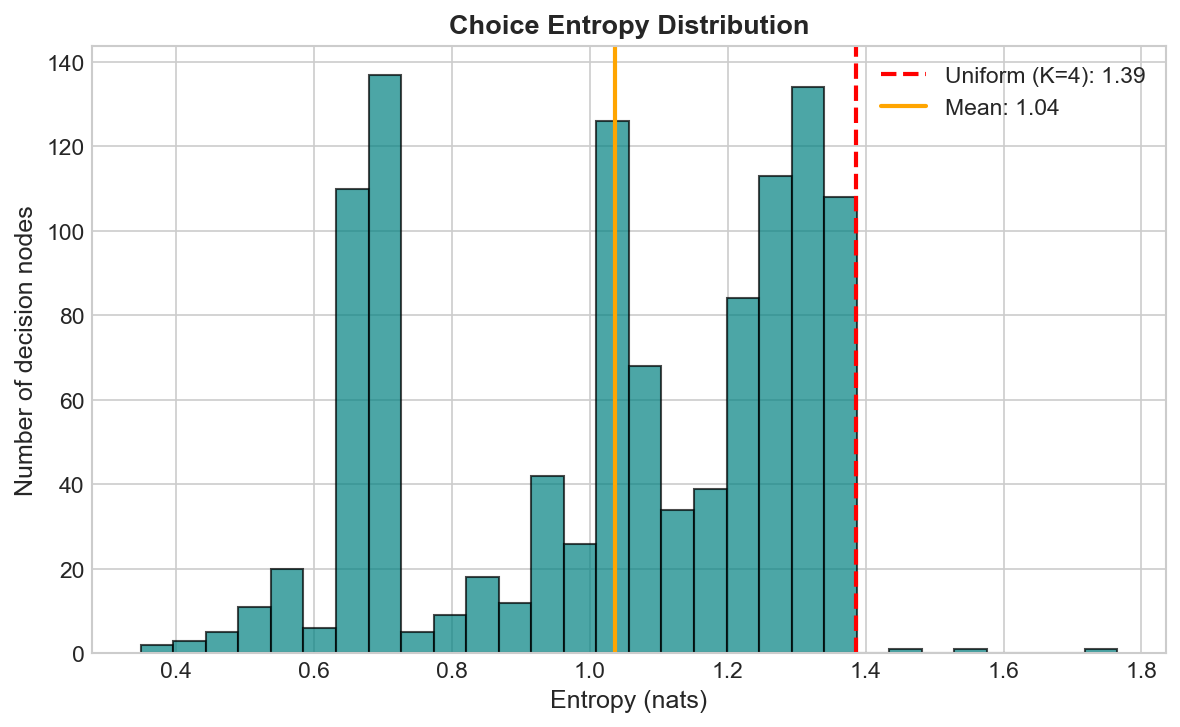}
    \caption{Distribution of user vote entropy across decision points. The dashed line marks the entropy of a uniform distribution over four choices, the typical candidate-set size. Mean observed entropy is lower than this baseline, indicating structured, non-random choice behavior without implying convergence to a single dominant outcome.}
    \label{fig:entropy}
\end{figure}
Figure~\ref{fig:teaser} provides an overview of the system used to collect the dataset. We created six AI-generated games for users to play. Each game establishes a context and plot, then asks players what should happen next at a branching point. Users select from a small set of options, typically four. The selected option is incorporated into the story, which continues to the next branching point. This process repeats until the end of the day at depth 4. Because the stories are open-ended, the system can generate additional days before concluding the narrative. During this voting process, we log the selected option, the alternatives, and additional metadata.

No payments or instruction-following tasks were required; users participated because the activity itself was enjoyable, yielding preferences that reflect natural, in situ behavior. Mean user-vote entropy is lower than the uniform four-choice baseline (Figure~\ref{fig:entropy}), indicating non-random preferences. This observation aligns with information-theoretic approaches to narrative evaluation, such as \textit{Fabula Entropy Indexing} \cite{castricato-etal-2021-fabula}, which posits that high-quality narratives exhibit low entropy in human question-answering tasks.

The current snapshot of our dataset consists of 44,226 preference votes by 8,167 unique users across the six games. We frame Rushes as a benchmark for predicting personalized user engagement in interactive narratives.

Our contributions are:
\begin{enumerate}
  \item A method for collecting large-scale, user-level preferences that reflect subjective engagement, including fun and interest;
  \item A large-scale dataset of 44,226 preference votes by 8,167 unique users across six narrative-based games (with multimodal content);
  \item A comprehensive benchmarking suite that establishes performance baselines using collaborative filtering and state-of-the-art LLMs, identifying an \textit{Engagement Gap} that challenges current alignment techniques;
  \item Code and prompts for the generation pipeline, together with an archived version of the environment, to be released publicly.
\end{enumerate}

We hope that our release will support researchers interested in personalized alignment, engagement modeling, and interactive narrative creation. Rushes is designed as a testbed for studying human preferences in open-ended, multimodal environments.

\section{Related Work}

\begin{table*}[t]
    \centering
    \small
    \resizebox{\textwidth}{!}{%
    \begin{tabular}{lccc}
        \toprule
        \textbf{Aspect} & \textbf{Rushes (Ours)} & \textbf{RLHF Chat Datasets} & \textbf{SeqRec Datasets} \\
        \midrule
        Multi-choice Decision Space ($k>2$) & \textbf{Yes} & Limited (Binary) & Yes \\
        Longitudinal User History & \textbf{Yes} & No (Stateless) & Yes \\
        Sequential Context Dependency & \textbf{Yes} & Limited & No (ID-based) \\
        Multimodal Assets (Image/Video) & \textbf{Yes} & No & Limited \\
        \midrule
        \textbf{Modeling Objective} & \textbf{Personalized Engagement} & \textbf{Safety \& Helpfulness} & \textbf{Clicks \& Purchase} \\
        \bottomrule
    \end{tabular}%
    }
    \caption{Comparison of \textit{Rushes} with prior work. Standard RLHF datasets focus on safety without longitudinal history, whereas sequential recommendation (SeqRec) datasets track item IDs rather than narrative context. \textit{Rushes} combines long-term user trajectories with rich, multimodal interactive narratives.}
    \label{tab:comparison}
\end{table*}

\paragraph{Interactive Narrative and Storytelling Benchmarks}

Recent interactive narrative benchmarks include \textit{TextQuests} \cite{phan2025textquestsgoodllmstextbased}, which uses classic interactive fiction to benchmark agents' reasoning and planning capabilities. Whereas \textit{TextQuests} evaluates whether an agent can solve a puzzle (competence), \textit{Rushes} evaluates whether a model can predict what a human wants to happen next (engagement). This distinction is important for developing agents that are not only capable but also enjoyable.

Similarly, \textit{What-If} \cite{huang2024whatifexploringbranchingnarratives} and \textit{Narrative Studio} \cite{ghaffari2025narrativestudiovisualnarrative} explore the generative mechanics of branching narratives. \textit{Narrative Studio}, for instance, employs Monte Carlo Tree Search (MCTS) to maximize narrative diversity during generation. \textit{Rushes} complements these system-focused works by providing data for evaluating the ``fun'' of the resulting generations. Although we employ similar semantic diversity checks in our generation pipeline (Section~\ref{sec:generate-games}) to prevent redundancy, our primary contribution is the capture of revealed human preferences within these diverse structures rather than the generation method itself.

\paragraph{Personalized Alignment and RLHF} Prior work on aligning large language models (LLMs) with human preferences has focused primarily on dimensions such as safety, helpfulness, or factual correctness \cite{ouyang2022training, bai2022training}. These datasets typically lack the longitudinal user history required for personalization.

Recent work has highlighted the ``cold-start'' problem in personalized alignment, arguing that static reward models fail to capture evolving user intent. \textit{LiteraryTaste} \cite{chung2025literarytastepreferencedatasetcreative} addresses this problem in creative writing, finding that explicit surveys (``stated preferences'') often fail to predict actual choices (``revealed preferences''). \textit{Rushes} captures revealed preferences through actions rather than surveys. Furthermore, \textit{LikeBench} \cite{rahman2025likebenchevaluatingsubjectivelikability} measures likability using simulated personas. \textit{Rushes} complements this work with human trajectories, whose preferences are often noisier and more context-dependent than those of simulated agents.

\paragraph{Drama Management and Interactive Narrative} 
Classical \textit{drama managers} (DMs) sought to adapt ongoing narratives to user preferences to maximize agency or enjoyment \cite{Yu_Riedl_2013,riedl2013interactive}. However, these systems often relied on handcrafted rules or symbolic planners, making them difficult to scale. Although neural approaches such as \textit{AI Dungeon} \cite{walton2019aidungeon} and \textit{Hierarchical Story Generation} \cite{fan-etal-2018-hierarchical} demonstrated the potential of open-ended text generation, they often lack the structured, longitudinal preference data necessary for personalized modeling. \textit{Rushes} modernizes this objective by scaling the environment with LLMs. Unlike classical DMs, which operate on restricted state spaces, \textit{Rushes} leverages the open-ended generation capabilities of frontier models while capturing revealed preferences at the scale of more than 44,000 interactions. These data can support user models for modern, LLM-based drama management.

\paragraph{Subjective Evaluation Metrics}

Measuring ``fun'' is difficult for standard reward models. \textit{WritingPreferenceBench} \cite{ying2025correctnessevaluatingsubjectivewriting} demonstrated that sequence-based reward models---the standard for RLHF---achieve only 52.7\% accuracy on subjective writing tasks, barely outperforming random chance. This result aligns with our finding that neural preference models struggle to beat popularity baselines in \textit{Rushes}. It suggests that modeling engagement may require architectural innovations, such as the \textit{Generative Reward Models} proposed by \citet{ying2025correctnessevaluatingsubjectivewriting}, which can reason about style and subtext.
\section{Rushes}
\label{sec:rushes-implementation}

\subsection{Game Generation}
\label{sec:generate-games}

\subsubsection{Generating branching narrative text}

In the current release, all narrative text and decision options are generated using GPT-4o with a temperature of 0.3 and a fixed prompting template (see Appendix~\ref{app:generation-pipeline} for all prompts). All generated nodes and options are stored prior to gameplay. Each story begins from a high-level synopsis that specifies the intended narrative trajectory, and the generation process recursively expands the story tree to a depth of four, yielding approximately 330 nodes that can be manually reviewed before release.

We selected a depth of four to mirror a concise narrative arc while keeping generation computationally manageable. Each decision node presents four options, a branching factor chosen to balance computational cost with sufficient variance to capture distinct behavioral strategies, such as aggressive, diplomatic, exploratory, or passive choices.
\paragraph{Semantic Diversity Enforcement} To prevent the generation of redundant options, a common failure mode in LLM storytelling, we employ a semantic similarity filter. At each decision node, an LLM-based checker compares the candidate option against previous options along the trajectory. If the option is judged too similar (considering action type, complexity, and narrative outcome), it is discarded and regenerated.

\paragraph{Lexical Diversity via Deterministic Paraphrasing} To mitigate lexical repetition and discourage users from navigating based on memorized surface text, we generate multiple semantic paraphrases for each option node during story generation. The number of variants scales with tree depth and expected traffic at each node, increasing lexical variety in high-traffic branches while limiting generation cost (see Appendix~\ref{app:paraphrase_derivation}). At runtime, the system selects a variant deterministically using a hash of the user's anonymized ID and the node identifier. This provides stable per-user lexical variation while preserving the underlying action semantics and avoiding real-time generation.

\subsubsection{Generating image, audio, and video}
Each narrative node is paired with multimodal assets to enhance immersion. Image generation is performed using a staged prompt construction pipeline (meta-prompts, similar to \citet{huang2024whatifexploringbranchingnarratives}) that improves character consistency and stylistic coherence. Images are then used as input to generative video models to create short clips using \textit{LTX-Video} \cite{hacohen2025ltxvideo}. Audio narration is synthesized using the \textit{Azure Text-to-Speech} (TTS) API with expressive styles.

\subsubsection{Generating narrative continuations}
To support multi-session narratives, Rushes enables dynamic story continuation across multiple "days" of gameplay. At the end of each day, we identify all active leaf nodes. We prune the exponential expansion by clustering leaf scenes into four broad narrative categories using an LLM. We then generate custom continuations for each active node that align with these categories.

\begin{table*}[t]
\centering
\small
\begin{tabular*}{\textwidth}{l@{\extracolsep{\fill}}ccccc|ccccc}
\toprule
 & \multicolumn{5}{c}{\textbf{Counts by Severity Level}} & \multicolumn{5}{c}{\textbf{Percentage (\%)}} \\
\cmidrule(lr){2-6} \cmidrule(lr){7-11}
\textbf{Dimension} & \textbf{0} & \textbf{2} & \textbf{4} & \textbf{6} & \textbf{Total $>$ 0} & \textbf{0} & \textbf{2} & \textbf{4} & \textbf{6} & \textbf{Total $>$ 0} \\
\midrule
Hate & 3,969 & 13 & 0 & 0 & 13 & 99.67 & 0.33 & 0.00 & 0.00 & 0.33 \\
Self-Harm & 3,943 & 31 & 8 & 0 & 39 & 99.02 & 0.78 & 0.20 & 0.00 & 0.98 \\
Sexual & 3,922 & 52 & 7 & 1 & 60 & 98.49 & 1.31 & 0.18 & 0.03 & 1.51 \\
Violence & 2,726 & 1,142 & 113 & 1 & 1,256 & 68.46 & 28.68 & 2.84 & 0.03 & 31.54 \\
\midrule
\textbf{Any Flag (non-zero)} & \textbf{2,674} & \textbf{1,182} & \textbf{124} & \textbf{2} & \textbf{1,308} & \textbf{67.15} & \textbf{29.68} & \textbf{3.11} & \textbf{0.05} & \textbf{32.85} \\
\bottomrule
\end{tabular*}
\caption{\textbf{Azure Content Safety analysis ($N=3{,}982$).} Distribution of safety severity scores across four dimensions. Severity levels range from 0 (safe) to 6 (high). The higher prevalence of low-severity violence flags reflects the action-adventure nature of the narrative genres.}
\label{tab:safety_stats}
\end{table*}

\subsubsection{Quality Control and Responsible AI}

All generated content is passed through an automated safety screening pipeline (Azure Content Safety API), with results summarized in Table~\ref{tab:safety_stats}. Across 3,982 screened generations, the system maintained strict safety standards on sensitive dimensions. The vast majority of content was classified as safe (Severity 0) for Hate (99.7\%), Sexual (98.5\%), and Self-Harm (99.0\%).

As expected for a dataset focused on action and adventure genres, the violence dimension had a higher flagging rate, with 31.5\% of generations scoring above Severity 0. Most flagged generations received Severity 2 (1,142 generations, or 28.7\% of all generations), consistent with standard genre tropes (e.g., science-fiction combat or dramatic tension) rather than graphic or gratuitous violence. Only 0.05\% of all generations received a Severity 6 score in any dimension.

All content was reviewed manually and approved by the authors. The gameplay interface also includes a user-facing reporting mechanism; however, we received no reports from users during the release.

\subsection{Analysis of Generated Games}

\subsubsection{Lexical and Semantic Diversity}
\begin{figure}[t]
    \centering
    \includegraphics[width=\columnwidth]{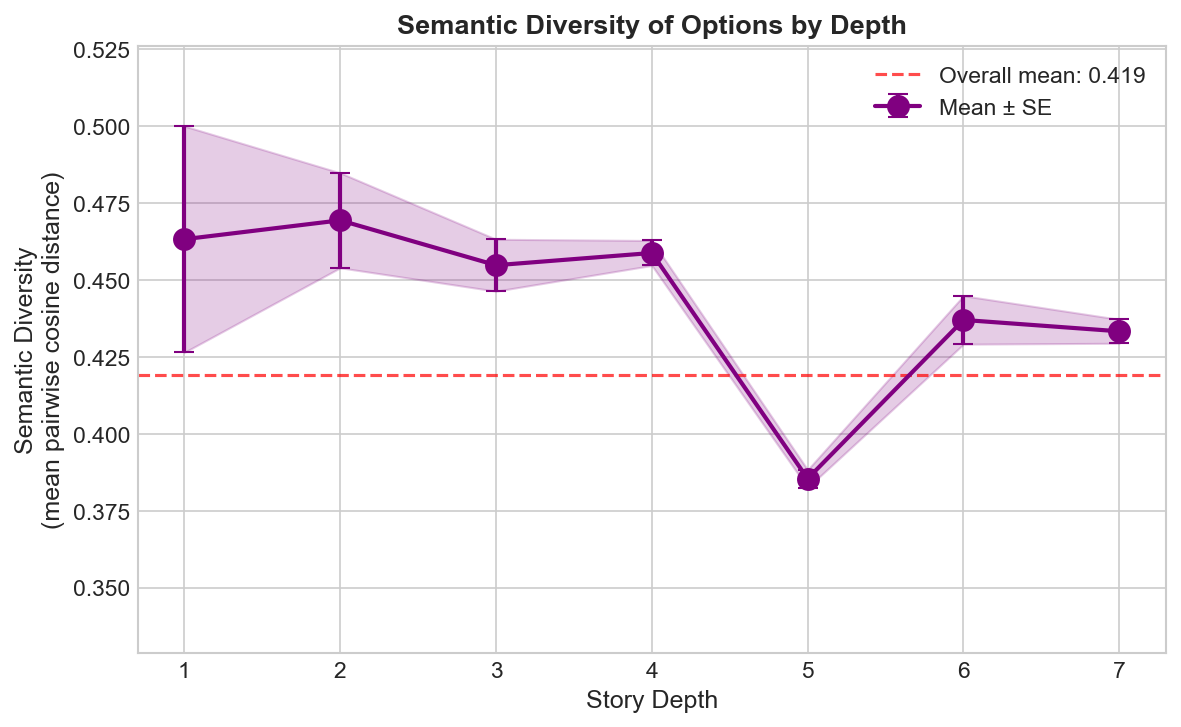}
    \caption{Mean semantic diversity, measured as pairwise cosine distance between embeddings from OpenAI's \textit{text-embedding-3-small} model. Error bars show standard error. The dip at depth 5 reflects end-of-day narrative consolidation.}
    \label{fig:semantic_diversity}
\end{figure}

We also evaluate the diversity of generated branches and options at each depth using average cosine distance in sentence-embedding space (Figure~\ref{fig:semantic_diversity}). Diversity drops at depth 5, reflecting the episodic generation pipeline: at the end of each day, leaf nodes are clustered into a small set of broad thematic continuations, temporarily consolidating the narrative state. This process helps maintain long-term coherence and manage complexity before the story expands into divergent paths on the subsequent day, mirroring serialized television. The variance in diversity also decreases over time, which may result from cumulative prompt growth constraining generation variability.

\subsubsection{Multimodal Asset Evaluation}
We observed occasional inconsistencies between text and generated media (images and video), and some users reported that narration quality varied across scenes. Despite these imperfections, the multimodal assets may have increased immersion by grounding decisions in a narrative world rather than isolated text prompts. This context may encourage in-world decision-making and reduce superficial text skimming. Because the benchmark evaluations in this paper are text-conditioned, we release the accompanying media primarily to preserve the context in which preferences were revealed and to support future multimodal modeling work.

\subsubsection{Summary}
Our goal in generating these narratives was not to use LLMs to break new ground in narrative construction, but rather to construct plausible stimuli that could be easily understood and enjoyed by our players.

These results suggest that users are presented, on average, with distinct and non-redundant alternatives at each decision point. This property is important for preference data collection: if options were trivially similar or repetitive, observed choices could be dominated by noise or superficial cues rather than substantive engagement.

Taken together with the low choice entropy observed in user behavior, the generation analysis supports the interpretation that Rushes captures structured, context-dependent decisions rather than arbitrary clicks. This validates the dataset as a suitable testbed for studying revealed preferences, sequential decision-making, and the limits of current alignment methods in interactive generative environments.

\subsection{Data Collection}
\paragraph{User Recruitment}

All participants in \textit{Rushes} were authenticated users recruited through the Xbox Insiders Program. Participation required signing in with verified Xbox credentials, providing persistent account-level identities rather than anonymous or crowdsourced accounts. Users voluntarily opted into the experience and engaged without financial incentives, reflecting intrinsic motivation and familiarity with interactive gaming environments.

\subsubsection{Logging and Schema}
Each click generates a vote, which is recorded in a standardized schema:
\begin{itemize}
    \item \texttt{user\_id} (anonymized identifier);
    \item \texttt{game\_id} and \texttt{level} (narrative depth);
    \item \texttt{vote} (selected option text) and \texttt{other\_options} (unselected candidates);
    \item Metadata: \texttt{time\_taken\_ms}, \texttt{user\_agent}, \texttt{session\_depth}.
\end{itemize}
Each record captures both the decision context and behavioral outcome, allowing reconstruction of complete narrative trajectories.

\subsubsection{Dataset Composition}
\begin{table}[t]
\centering
\small
\resizebox{\columnwidth}{!}{%
\begin{tabular}{l r}
\toprule
Statistic & Value \\
\midrule
Active users & 8{,}167 \\
Total decision events (votes) & 44{,}226 \\
Number of games & 6 \\
Average trajectory depth (decisions per playthrough) & 5.4 \\
Average games played per user & 1.4 \\
Users who played all 6 games & 195 \\
Typical candidate set size per decision & 4 options \\
\bottomrule
\end{tabular}%
}
\caption{Summary statistics for the current Rushes snapshot. Each decision event logs the full candidate set and the user’s chosen option, plus metadata (e.g., time taken and session depth).}
\label{tab:rushes-dataset-stats}
\end{table}

The final dataset comprises 44,226 distinct decision events generated by 8,167 unique users across the six available titles (Table~\ref{tab:rushes-dataset-stats}). The distribution of user engagement follows a long-tailed pattern typical of gaming environments. While the average participant interacted with 1.4 games, a dedicated core of 195 ``power users'' engaged with all six narrative environments.

In terms of session length, the average trajectory reached a depth of 5.4 decision points. Because the standard ``day'' cycle concludes at depth 4, this indicates that many users continued past the initial narrative loop to experience multi-day continuations. The participant pool consists exclusively of authenticated Xbox Insiders, and the interface was presented in English. Recruitment therefore likely favored users familiar with English-language branching game mechanics.

We also observe that engagement is non-uniform. As shown in Figure~\ref{fig:entropy}, mean user-vote entropy (1.04 nats) is lower than the uniform four-choice baseline (1.39 nats), indicating structured, context-dependent narrative preferences at the aggregate level.

\subsubsection{Preference Transformation and Modeling}
To support alignment research, the raw interaction logs can be transformed into training-ready formats. We convert each ``choose 1 of $k$'' decision (typically $k=4$) into $k-1$ distinct pairwise comparisons, denoted as $(o_{chosen} \succ o_{rejected})$. These pairs encode the observed choice as a preference for the selected option over each alternative, enabling the training of standard reward models and Direct Preference Optimization (DPO).

\section{Experiments and Results}
\label{sec:benchmark}

\begin{table*}[t]
    \centering
    \small
    \resizebox{\textwidth}{!}{%
    \begin{tabular}{lccl} 
    \toprule
    \textbf{Baseline} & \textbf{Accuracy [95\% CI]} & \textbf{\textit{N} Samples} & \textbf{Description} \\
    \midrule
    \textbf{SVD Collab. Filtering} & \textbf{0.3773 [0.3669, 0.3878]} & 8293 & Matrix factorization on user-option interactions \\
    Popularity (Most Freq) & 0.3639 [0.3536, 0.3743] & 8293 & Always select the historically most popular option \\
    GPT-5 with History & 0.3423 [0.3321, 0.3526] & 8293 & GPT-5 prompted with user history \\
    SASRec & 0.3406 [0.3304, 0.3509] & 8293 & Self-Attentive Sequential Recommendation \\
    Semantic Classifier & 0.3000 [0.2902, 0.3100] & 8293 & Fine-tuned DeBERTa-v3 on context+option pairs \\
    Random (Uniform) & 0.2541 [0.2448, 0.2636] & 8293 & Uniform selection among 4 available options \\
    \bottomrule
    \end{tabular}%
    }
    \caption{Main baseline accuracy on the \textit{Rushes} test set with 95\% Wilson confidence intervals. All models are evaluated on the same held-out test split.}
    \label{tab:main_results}
\end{table*}

\begin{table}[t]
    \centering
    \small
    \resizebox{\columnwidth}{!}{%
    \begin{tabular}{lcc}
    \toprule
    \textbf{Depth} & \textbf{Accuracy [95\% CI]} & \textbf{\textit{N} Samples} \\
    \midrule
    0 & 0.3048 [0.2726, 0.3390] & 735 \\
    1 & 0.3100 [0.2854, 0.3357] & 1300 \\
    2 & 0.3740 [0.3494, 0.3993] & 1441 \\
    3 & 0.3763 [0.3620, 0.3908] & 4361 \\
    4 & \textbf{0.4207 [0.3761, 0.4666]} & 454 \\
    \bottomrule
    \end{tabular}%
    }
    \caption{Popularity baseline accuracy by narrative depth. Accuracy peaks at depth 4. Two test events without matched narrative-depth metadata are omitted.}
    \label{tab:ablation_depth}
\end{table}

\begin{table}[t]
    \centering
    \small
    \resizebox{\columnwidth}{!}{%
    \begin{tabular}{lcc}
    \toprule
    \textbf{History Type} & \textbf{Accuracy [95\% CI]} & \textbf{\textit{N} Samples} \\
    \midrule
    Same-game history & \textbf{0.3886 [0.3775, 0.3999]} & 7310 \\
    Cross-game history & 0.2909 [0.2634, 0.3201] & 983 \\
    All history (SVD) & 0.3773 [0.3669, 0.3878] & 8293 \\
    \bottomrule
    \end{tabular}%
    }
    \caption{Impact of history source on prediction accuracy. Same-game history is more predictive than cross-game history.}
    \label{tab:ablation_history}
\end{table}

\begin{table}[t]
    \centering
    \small
    \resizebox{\columnwidth}{!}{%
    \begin{tabular}{lcc}
    \toprule
    \textbf{Method} & \textbf{Sparse Players [95\% CI]} & \textbf{Active Players [95\% CI]} \\
    \midrule
    Random & 0.2540 [0.2450, 0.2632] & 0.2491 [0.2008, 0.3045] \\
    Popularity & 0.3523 [0.3390, 0.3660] & \textbf{0.3778 [0.3618, 0.3940]} \\
    SVD & \textbf{0.3840 [0.3704, 0.3978]} & 0.3683 [0.3523, 0.3845] \\
    \bottomrule
    \end{tabular}%
    }
    \caption{Accuracy stratified by user activity level. Sparse players played one game; active players played two or more games. Bold indicates the highest accuracy in each group.}
    \label{tab:new_vs_returning}
\end{table}

\begin{table}[t]
\centering
\small
\resizebox{\columnwidth}{!}{%
\begin{tabular}{l c c}
\toprule
\textbf{Model} & \textbf{Accuracy [95\% CI]} & \textbf{N Samples} \\
\midrule
GPT-4o (Zero-Shot) & 0.3030 [0.2931, 0.3130] & 8293 \\
GPT-5 (Zero-Shot) & 0.3090 [0.2991, 0.3191] & 8293 \\
\midrule
GPT-4o (with History) & 0.3390 [0.3288, 0.3493] & 8293 \\
GPT-5 (with History) & \textbf{0.3423 [0.3321, 0.3526]} & 8293 \\
\bottomrule
\end{tabular}%
}
\caption{Impact of model scaling and context. Scaling from GPT-4o to GPT-5 yields marginal gains ($<1\%$). Adding historical context provides a larger boost ($\approx 4\%$).}
\label{tab:frontier_models}
\end{table}

\paragraph{Task Definition}
We frame evaluation as event-level, text-based candidate-choice prediction. At each decision point, the model observes the narrative context, the available candidate options, and the user's interaction history up to that point and must predict which single option the user selected.

We use top-1 accuracy because \textit{Rushes} captures single, irreversible user decisions rather than graded preferences or ranked lists. Pairwise and ranking metrics would answer a different question by decomposing one holistic choice into multiple comparisons, obscuring the difficulty of predicting the user's committed action.

\paragraph{Evaluation Protocol}
We evaluate event-level top-1 choice prediction using a user-stratified chronological split. For each user, interactions are ordered by time, with the first 80\% used for training and the remaining 20\% held out for testing, ensuring that all test decisions occur after the user's training history.

\subsection{Main Results}

\paragraph{Popularity Bias as a Strong Baseline} We observe that SVD (37.73\%) slightly outperforms the popularity baseline (36.39\%), suggesting that \textit{Rushes} contains personalized signals that distinguish individual users from the aggregate mean. However, frontier LLMs still fail to capture this signal, falling behind both classical collaborative filtering and simple popularity heuristics. This result mirrors findings in recommender systems, where popularity bias can overshadow user-specific signals, and in recent creative-writing benchmarks~\citep{ying2025correctnessevaluatingsubjectivewriting,chung2025literarytastepreferencedatasetcreative}. In these subjective domains, standard reward models frequently struggle to decouple ``generic quality'' from ``personal appeal,'' defaulting to safe, high-probability tokens rather than riskier, context-dependent predictions.

We further evaluate \textit{SASRec} (Self-Attentive Sequential Recommendation) \cite{kang2018sasrec} to test whether specialized sequential modeling can bridge the engagement gap. \textit{SASRec} achieves 34.06\% accuracy, performing on par with the much larger GPT-5 with history (34.23\%). However, both methods fail to outperform the popularity baseline (36.39\%) and trail SVD (37.73\%). This result suggests that, in the current formulation, sequential attention alone does not outperform simpler identity-based baselines.

\subsection{Ablation Studies}
To examine whether user preferences contain personalized signal beyond global trends, we analyze the limits of the popularity baseline. Popularity reaches 36.4\% accuracy, leaving 63.6\% of choices uncaptured by a global majority heuristic. This residual reflects substantial heterogeneity across decisions, but it does not by itself distinguish stable user-specific preferences from context-dependent variation. SVD's improvement over popularity provides more direct evidence of personalized signal.

\subsubsection{Engagement by Narrative Depth}
We analyzed the popularity baseline's accuracy at different depths of the story tree (Table~\ref{tab:ablation_depth}). Accuracy consistently rises as the narrative progresses, peaking at depth 4 (42.07\%). This suggests that as users deepen their engagement with a specific narrative arc, their choices become easier to predict under popularity heuristics.

\subsubsection{The Role of History}
We evaluated how user history impacts prediction in Table~\ref{tab:ablation_history}.

\paragraph{Same-Game History:} When a user has history within the current game, accuracy is 38.86\%.

\paragraph{Cross-Game History:} When a user has history only from different games, accuracy drops to 29.09\%. This 9.8-point gap suggests that preferences are highly context-dependent. A user's preference for ``action'' in a science-fiction game does not necessarily transfer to a mystery game, highlighting the difficulty of transfer learning in narrative engagement.

\subsubsection{Active vs. Sparse Players}
As shown in Table~\ref{tab:new_vs_returning}, active players (those returning for two or more games) are harder for SVD to predict (36.83\%) than sparse players (38.40\%). The popularity baseline performs best for active players (37.78\%). One possible explanation is that highly engaged users may actively explore the system, making choices that deviate from their own history while aligning with globally interesting content. We leave disentangling exploratory behavior from model limitations to future work.

\subsubsection{Frontier Model Scaling: GPT-5 vs. GPT-4o}
To assess whether reasoning capabilities improve alignment, we evaluated GPT-5 against GPT-4o on the full test set (Table~\ref{tab:frontier_models}). Scaling offers marginal zero-shot gains (30.9\% for GPT-5 vs.\ 30.3\% for GPT-4o), whereas adding user history provides a larger boost, lifting GPT-5 to 34.2\%. However, GPT-5 with the available user history still fails to outperform the popularity baseline (36.4\%). This result reinforces the finding from \textit{WritingPreferenceBench} that scaling model capacity or context alone does not fully bridge the engagement gap. Capturing ``fun'' may require explicit alignment with subjective values and idiosyncratic preferences that diverge from population trends.

\section{Conclusion}

As large language models evolve from passive tools to interactive agents, modeling engagement becomes as important as modeling competence. \textit{Rushes} shows that organic user choices exhibit structured, non-random patterns that remain difficult for current frontier LLMs to predict under standard training and evaluation paradigms. The performance gap between a simple personalized-history model and state-of-the-art LLMs highlights the difficulty of modeling engagement in sequential narrative settings. \textit{Rushes} provides a diagnostic benchmark for studying these limitations and advancing research on pluralistic alignment, in which models must adapt to diverse, subjective notions of meaningful experiences rather than converge to population-level averages.

\section*{Ethical considerations}

\textbf{Data Provenance and Recruitment} Participants were recruited through the Xbox Insiders Program (Public Ring), a platform where users voluntarily opt-in to test pre-release content and experiments. Users were presented with a clear consent page explaining that their anonymized interaction data would be logged for research purposes and potentially released as an open-source dataset. Participation was strictly voluntary, and no financial incentives were provided; users engaged with the system solely for the intrinsic value of the gameplay experience.

\textbf{Responsible AI and Dual Use} We release the Rushes dataset and the associated code to foster research into personalized alignment. However, we acknowledge that methods for optimizing "engagement" can be dual-use, potentially applicable to addictive design patterns or dark patterns in UI/UX. We condemn the use of this dataset for manipulative purposes and urge the community to focus on pluralistic alignment—serving diverse user needs—rather than engagement maximization for its own sake. The release is governed by a license that prohibits malicious use, and no personal identifiable information (PII) is included in the release; all user IDs have been cryptographically hashed.

\section*{Limitations}

This report relates to Rushes as implemented using GPT-4o. The results shown in the demonstration will differ if other LLMs are used. No claim is made to the superiority of performance of any LLM. Outputs will vary under different temperature settings and with different prompting strategies and formats. 

This system relates to games generation only. In principle, the approach taken by Rushes should be extensible to multimodal games generation, particularly those with a visual component, e.g., in a storyboarding application, but that is beyond the scope of this work. 

The system is implemented using English-language prompts. It has not been investigated in other languages. Given our observation that Rushes appears to perform better on better documented settings, we expect that some degradation may occur when used with languages other than English. 

As we have noted elsewhere, the architecture of this system readily lends itself to iterative editing and reprompting for further exploration of paths. Full implementation of this feature, however, involves application-specific considerations and harm mitigations for public presentation. This must be left for future work. 

Given the recruitment platform, the user base is demographically skewed towards gaming-literate populations who are likely comfortable with branching narrative mechanics. Furthermore, as the generated content and interface were presented exclusively in English, the dataset reflects the preferences of English-speaking users, predominantly from regions with high Xbox Insider adoption. Consequently, the engagement patterns observed in Rushes should not be interpreted as a universal baseline for human preference but rather as a specific reflection of this gamer-centric demographic. We explicitly caution against generalizing these findings to non-gaming or non-English speaking populations without further validation.

\section*{Acknowledgments}
We would like to thank Leland Olney for his instrumental support and partnership in facilitating the Xbox Insiders recruitment and data collection process. We are also deeply grateful to Chris Brockett for his insightful feedback and support throughout the development of this project.

\bibliography{custom}

\appendix

\onecolumn
\raggedbottom
\section{Rushes Game Generation Pipeline}
\label{app:generation-pipeline}

\subsection{Configuration Parameters}

\begin{table}[H]
\centering
\begin{tabular}{|l|l|}
\hline
\textbf{Parameter} & \textbf{Value} \\
\hline
LLM Model & GPT-4o \\
API Version & 2024-10-01-preview \\
Expected Players & 5,000 \\
Options per Level & 4 \\
Maximum Depth & 4 levels per day \\
Speech Service & Azure TTS (Fable HD) \\
Image Model & FLUX.1 schnell \\
\hline
\end{tabular}
\caption{System configuration parameters}
\end{table}

\subsection{Deriving the paraphrase scaling rule}
\label{app:paraphrase_derivation}

Rushes pre-generates multiple surface realizations (paraphrases) for each option to reduce repeated wording across users. Let $P$ be the expected number of players for a game, $b$ the branching factor (number of options per node; in our setup $b=4$), and $d$ the depth index of a decision node (root at $d=0$).

Assuming players are approximately evenly distributed across branches,\footnote{This assumption is used only to size the paraphrase budget; the actual distribution may be skewed.} the expected number of players who reach a particular node at depth $d$ is:
\begin{equation}
\mathbb{E}[\#\text{players at node depth } d] \approx \frac{P}{b^{d}}.
\end{equation}
Each such node presents $b$ options. Under the same uniformity assumption, let $M(d)$ denote the expected number of players who select a particular option at depth $d$:
\begin{equation}
M(d) = \mathbb{E}[\#\text{selections per option at depth } d]
     \approx \frac{P}{b^{d+1}}.
\end{equation}
Allocating one surface variant per expected selection would grow linearly with player traffic and be prohibitively expensive. Instead, \textit{Rushes} uses a square-root heuristic. Let $K(d)$ denote the total number of surface variants available for an option at depth $d$, including the original phrasing:
\begin{equation}
K(d) = \left\lceil \sqrt{M(d)} \right\rceil
     = \left\lceil \sqrt{\frac{P}{b^{d+1}}} \right\rceil.
\end{equation}
Since we store the original phrasing plus $V(d)$ additional paraphrases, $K(d)=V(d)+1$, yielding:
\begin{equation}
V(d) = \left\lceil \sqrt{\frac{P}{b^{d+1}}} \right\rceil - 1.
\end{equation}
This heuristic increases lexical variety with expected traffic while keeping generation costs sublinear. It reduces repeated wording but does not guarantee a unique variant for every player.

\paragraph{Variant assignment.}
At interaction time, a single variant is selected deterministically using a hash of (anonymized) \texttt{user\_id} and the \texttt{(node\_id, option\_id)} pair. This provides stable per-user lexical variation without any on-demand generation.

\subsection{Main Generation Pipeline}
\begin{mdframed}[linewidth=0.5pt, nobreak=true]
\noindent\textbf{GenerateNewGame: Create Complete Interactive Narrative}
\begin{algorithmic}[1]
\REQUIRE $synopsis$, $game\_name$, $num\_options$, $max\_depth$
\ENSURE $game\_uuid$, complete game data
\STATE $game\_uuid \gets$ GenerateUUID()
\STATE $setup \gets$ LLM + StorySetup($synopsis$)
\STATE $theme \gets setup.theme$ \COMMENT{Visual themes for consistency}
\STATE $results \gets$ LLM + CreateStory(
\STATE \hspace{2em} $synopsis$, $num\_options$, $max\_depth$, 
\STATE \hspace{2em} $levels$, $checkpoint\_file$)
\STATE SaveToFile($game\_name$, $results.levels$, $theme$)
\RETURN $game\_uuid$
\end{algorithmic}
\end{mdframed}

\subsection{Story Setup and Theme Generation}

\subsubsection{Theme Extraction Prompt}
\begin{promptbox}
\textbf{System Prompt:}
\begin{verbatim}
TASK: Storywriting
INSTRUCTIONS: You are a writer tasked with creating visuals 
for a short story based on a provided synopsis. Give the user 
a concise but detailed description of the overall art style 
of the story and look of the subjects.

For the medium, specify: digital art, illustration, oil painting, 
3D rendering, photography, etc.
For the style, specify: impressionist, surrealist, pop art, 
realism, fantasy, etc.
For the colors, list the main colors that should be used.
For lighting, specify: natural, artificial, neon, dark, bright, etc.
Include additional details using EXTRA that would help an artist.
Use only keywords and short phrases. End with 'END'.

EXAMPLE:
Synopsis: A detective investigates mysterious disappearances 
in dystopian futuristic America.
OUTPUT:
MEDIUM: Digital art
ARTISTIC STYLE: hyperrealistic, fantasy, dark art
COLORS: iridescent gold, deep purple, midnight blue
LIGHTING: studio lighting, shadows at sharp angles
EXTRA: sci-fi elements, neon lighting, retro-futuristic tech
END
\end{verbatim}

\textbf{User Input:} Synopsis: \{synopsis\}

\textbf{Assistant Output:} \{theme\}
\end{promptbox}

\subsection{Recursive Story Generation}
\begin{mdframed}[linewidth=0.5pt, nobreak=true]
\noindent\textbf{CreateStory: Generate Branching Narrative Tree}
\begin{algorithmic}[1]
\REQUIRE $synopsis$, $n\_options$, $max\_depth$, $levels$
\ENSURE Complete story tree with multiple paths
\STATE \textbf{System:} Set context as game design expert
\STATE \textbf{User:} "I want a story about \{synopsis\}. Begin writing and stop at CROSSROADS."
\STATE \textbf{Assistant:} $initial\_story \gets$ LLM.generate(stop="CROSSROADS")
\STATE $levels[$"start"$] \gets \{$
\STATE \hspace{2em} dialog: [$initial\_story$],
\STATE \hspace{2em} depth: 1,
\STATE \hspace{2em} menu: \{buttons: []\}
\STATE \}
\STATE $levels \gets$ GenerateLevel(
\STATE \hspace{2em} $initial\_story$, depth=0, max\_depth=$max\_depth$,
\STATE \hspace{2em} $n\_options$, level\_id="start", $checkpoint\_file$)
\RETURN $levels$
\end{algorithmic}
\end{mdframed}

\subsection{Level Generation with Branching}
\begin{mdframed}[linewidth=0.5pt, nobreak=true]
\noindent\textbf{GenerateLevel: Create Single Story Node with Options}
\begin{algorithmic}[1]
\REQUIRE $story$, $depth$, $max\_depth$, $n\_options$, $level\_id$
\ENSURE Updated levels with new branches
\STATE Create level entry in $levels[level\_id]$ with $story$, $depth$
\IF{$depth \geq max\_depth$}
    \RETURN $levels$ \COMMENT{Reached maximum depth}
\ENDIF
\STATE $n\_variations \gets \lceil\sqrt{EXPECTED\_PLAYERS / n\_options^{depth+1}}\rceil - 1$
\STATE $options \gets$ LLM + CreateOptions($n\_options$, $depth$, $n\_variations$)
\STATE $parent\_menu\_texts \gets$ Extract titles from $options$
\STATE $seen\_options \gets$ Accumulate seen options for uniqueness checking
\FOR{each $new\_level\_id$, $option$ in $options$}
    \STATE Ensure $new\_level\_id$ is unique (append counter if needed)
    \STATE \textbf{User:} "User chose: \{option.action\}"
    \IF{$depth = max\_depth - 1$}
        \STATE \textbf{User:} "This is the last level. Provide conclusion. ENDSTORY."
    \ELSE
        \STATE \textbf{User:} "Continue story, stop at next CROSSROADS."
    \ENDIF
    \STATE \textbf{Assistant:} $option\_story \gets$ LLM.generate(stop=["CROSSROADS", "ENDSTORY"])
    \STATE $levels[new\_level\_id] \gets$ Create new level with $option\_story$
    \STATE $levels \gets$ GenerateLevel(
    \STATE \hspace{2em} $option\_story$, $depth+1$, $max\_depth$, 
    \STATE \hspace{2em} $n\_options$, $new\_level\_id$)
\ENDFOR
\RETURN $levels$
\end{algorithmic}
\end{mdframed}

\subsection{Option Generation with Variations}

\subsubsection{Option Creation Prompt}
\begin{promptbox}
\textbf{User Prompt:}
\begin{verbatim}
Provide {n_options} short and descriptive options on what 
the user could do next.

REQUIREMENTS:
- Each choice must be fully ACTIONABLE (not vague or mental)
- Each choice must be narratively and visually engaging
- Each choice must be unique in this level
- Titles must be in lowercase snake_case format

NARRATIVE GUIDANCE:
{depth > 0: "Unfold narrative smoothly while introducing 
action-heavy, tense, and cinematic events."
else: "Since we're at the beginning, unfold smoothly with 
actionable options without building tension yet."}

OUTPUT FORMAT:
<think>[Your reasoning for each option]</think>
OPTION 1: [title]: [option description]
OPTION 2: [title]: [option description]
...
OPTION {n_options}: [title]: [option description]
ENDOPTIONS
\end{verbatim}
\end{promptbox}

\begin{mdframed}[linewidth=0.5pt, nobreak=true]
\noindent\textbf{CreateOptions: Generate Diverse Action Choices}
\begin{algorithmic}[1]
\REQUIRE $n\_options$, $depth$, $enforce\_unique$, $n\_variations$
\ENSURE Set of unique, actionable options with variations
\STATE $options \gets$ Empty dictionary
\WHILE{len($options$) $< n\_options$}
    \STATE \textbf{User:} Request $n\_options$ using format above
    \STATE \textbf{Assistant:} $response \gets$ LLM.generate(stop="ENDOPTIONS")
    \STATE $parsed\_options \gets$ ExtractOptions($response$) via regex
    \FOR{each $option$ in $parsed\_options$}
        \IF{$enforce\_unique$}
            \STATE $is\_similar \gets$ CheckSimilarity($option.text$, $seen\_options$)
            \IF{$is\_similar$}
                \STATE Continue \COMMENT{Skip similar option}
            \ENDIF
        \ENDIF
        \IF{$n\_variations > 0$}
            \STATE $expanded \gets$ ExpandOption($option$, $n\_variations$)
            \STATE $option.variations \gets expanded.variations$
            \STATE $option.details \gets expanded.details$
            \STATE $option.outcome \gets expanded.outcome$
        \ENDIF
        \STATE Add $option$ to $options$
    \ENDFOR
\ENDWHILE
\RETURN $options$
\end{algorithmic}
\end{mdframed}

\subsection{Similarity Checking for Uniqueness}
\begin{promptbox}
\textbf{System Prompt:}
\begin{verbatim}
You are a story similarity checker.
Task: Determine if the provided option is overly similar to 
any previous nodes that have been seen by the user.

Analyze similarity across these dimensions:
- Nature of Action (combat vs. dialogue vs. exploration)
- Complexity (simple vs. multi-step)
- Physicality (physical action vs. mental/social)
- Outcome (consequences and story progression)

OUTPUT FORMAT:
<think>[Your detailed analysis comparing current option 
to previous nodes]</think>
<answer>[True or False: True ONLY if current option is 
overly similar to a previous node, otherwise False]</answer>
ENDRESPONSE
\end{verbatim}

\textbf{User Input:} 
\begin{verbatim}
Previous nodes: {seen_options}
Current option: {current_option_text}
\end{verbatim}

\textbf{Assistant Output:} \{analysis + answer\}
\end{promptbox}

\subsection{Option Expansion for Variation}
\begin{mdframed}[linewidth=0.5pt, nobreak=true]
\noindent\textbf{ExpandOption: Generate Detailed Variations}
\begin{algorithmic}[1]
\REQUIRE $option$, $n\_variations$
\ENSURE Expanded option with process details and outcome
\STATE \textbf{System:} "Generate concrete description of option and outcome."
\STATE \textbf{User:} "Option Title: \{option[0]\}$\backslash$nOption Action: \{option[1]\}"
\STATE \textbf{Assistant:} $details\_response \gets$ LLM.generate(
\STATE \hspace{2em} format="DETAILS: Details/Process: ... Immediate Outcome: ...")
\STATE $details \gets$ Extract from $details\_response$
\STATE $outcome \gets$ Extract from $details\_response$
\STATE \textbf{System:} "Generate \{n\_variations\} variations of Details/Process"
\STATE \hspace{2em} "Keep title, action, outcome same. Vary only process."
\STATE \textbf{Assistant:} $variations\_response \gets$ LLM.generate(
\STATE \hspace{2em} format="VARIATION X: Details/Process: ...")
\STATE $variations \gets$ Extract all variations via regex
\RETURN \{details, outcome, variations\}
\end{algorithmic}
\end{mdframed}

\begin{promptbox}
\textbf{Paraphrase Generation Prompt:}
\begin{verbatim}
TASK: Generate Variations of Options
Given the current option, generate {n_variations} distinct, 
actionable variations, keeping the general idea consistent.

RULES:
1. Preserve every piece of context from the original:
   - Character names, locations, roles, relationships
   - Specializations or backstory
2. Each variation must:
   - Begin by restating essential context
   - Offer fresh style or approach in two sentences
   - Avoid repeating exact wording while keeping details
   - Not assume prior knowledge

FORMAT:
VARIATION X:
Option Action: [brief two sentence description]

Stop when you have exactly {n_variations} variations.
Print ENDOPTIONS.
\end{verbatim}
\end{promptbox}

\subsection{Game Continuation Algorithm}

For multi-day games, the system continues stories from active player paths:
\begin{mdframed}[linewidth=0.5pt, nobreak=true]
\noindent\textbf{ContinueGame: Extend Game from Active Storylines}
\begin{algorithmic}[1]
\REQUIRE $game\_uuid$, $current\_day$, $n\_storylines$, $num\_options$
\ENSURE New day's story branches
\STATE $game\_data \gets$ LoadFromDatabase($game\_uuid$)
\STATE $levels \gets$ LoadFromDatabase($game\_uuid$, $current\_day$)
\STATE $story\_tree \gets$ GenerateStoryTree($levels$, root="start")
\STATE $current\_depth \gets 5$ \COMMENT{End of previous day}
\STATE $storylines \gets$ GetActiveStorylines($votes\_db$, $game\_uuid$, $current\_depth$)
\STATE $stories \gets$ Map storylines to story text from $story\_tree$
\IF{len($stories$) = 0}
    \RETURN \COMMENT{No active players}
\ENDIF
\STATE $merged \gets$ LLM + MergeOptions(
\STATE \hspace{2em} $n\_storylines$, $num\_options$, $stories$, $levels$, $current\_depth$)
\STATE $next\_day \gets current\_day + 1$
\STATE $results \gets$ LLM + ContinueStory(
\STATE \hspace{2em} $synopsis$, $num\_options$, $current\_depth+2$, $current\_depth$,
\STATE \hspace{2em} $merged$, $story\_tree$, $levels$)
\STATE SaveToDatabase($game\_uuid$, $next\_day$, $results.levels$)
\RETURN $results$
\end{algorithmic}
\end{mdframed}

\subsection{Image Prompt Generation}

\subsubsection{Character Extraction and Management}
\begin{mdframed}[linewidth=0.5pt, nobreak=true]
\noindent\textbf{GenerateImagePrompt: Create Stable Diffusion Prompts}
\begin{algorithmic}[1]
\REQUIRE $themes$, $caption$, $subjects$
\ENSURE Image prompt for scene
\STATE \textbf{System:} "Extract characters from text. Convert to snake\_case."
\STATE \hspace{2em} "Use existing names if already in EXISTING SUBJECTS."
\STATE \textbf{User:} "EXISTING SUBJECTS: \{subjects.keys()\}$\backslash$nINPUT: \{caption\}"
\STATE \textbf{Assistant:} $scene\_subjects \gets$ LLM.generate(format="char1, char2 ENDOUTPUT")
\STATE Parse $scene\_subjects$ into list
\FOR{each $subject$ in $scene\_subjects$}
    \IF{$subject$ not in $subjects$}
        \STATE \textbf{System:} "Create detailed character description."
        \STATE \hspace{2em} "Format: species, gender, age, appearance, clothing, traits"
        \STATE \hspace{2em} "Must be fully clothed and appropriate."
        \STATE \textbf{User:} "Create character named: \{subject\}"
        \STATE \textbf{Assistant:} $char\_desc \gets$ LLM.generate(stop="END")
        \STATE $subjects[subject] \gets char\_desc$
    \ENDIF
\ENDFOR
\STATE \textbf{System:} "Create detailed Stable Diffusion prompt."
\STATE \hspace{2em} "Third-person, vivid visual details, comma-separated."
\STATE \hspace{2em} "Match themes: \{themes\}"
\STATE \hspace{2em} "AVAILABLE CHARACTERS: \{subjects for scene\_subjects\}"
\STATE \textbf{User:} "I want an image about: '\{caption\}'"
\STATE \textbf{Assistant:} $image\_prompt \gets$ LLM.generate()
\RETURN $image\_prompt$, $subjects$
\end{algorithmic}
\end{mdframed}

\subsection{Audio Generation with SSML}
\begin{promptbox}
\textbf{Audio Prompt Generation:}
\begin{verbatim}
TASK: Audio Synthesis
You are an expert in generating audio prompts for text.
Generate SSML to narrate the scene, including character 
dialogue in distinct voices.

- Use voice `en-US-FableMultilingualHD` for narration
- Set mstts:express-as style to `narration-professional`
- Use <prosody> for rate, pitch, volume adjustments
- Use <mstts:express-as> for character roles and styles

OUTPUT FORMAT: <speak>SSML Prompt</speak>

INPUT: {caption}
\end{verbatim}
\end{promptbox}

\subsection{Media Generation Pipeline}
\begin{mdframed}[linewidth=0.5pt, nobreak=true]
\noindent\textbf{GenerateGameMedia: Create Images and Audio}
\begin{algorithmic}[1]
\REQUIRE $game\_uuid$, $day$
\ENSURE Image prompts and audio files
\STATE $levels \gets$ LoadFromDatabase($game\_uuid$, $day$)
\STATE $game\_data \gets$ LoadFromDatabase($game\_uuid$)
\TRY
    \STATE $images\_data \gets$ LoadFromDatabase($game\_uuid$, $day$, type="images")
    \STATE $theme \gets images\_data.theme$
    \STATE $subjects \gets images\_data.subjects$
\CATCH
    \STATE $setup \gets$ LLM + StorySetup($synopsis$)
    \STATE $theme \gets setup.theme$
    \STATE $subjects \gets \{\}$
\ENDTRY
\STATE $dialog\_texts \gets$ Extract dialog from all levels
\FOR{each $level\_id$, $text$ in $dialog\_texts$}
    \IF{$level\_id$ not in $image\_prompts$}
        \STATE $prompt \gets$ LLM + GenerateImagePrompt($theme$, $text$, $subjects$)
        \STATE $image\_prompts[level\_id] \gets prompt.image\_prompt$
        \STATE $subjects \gets prompt.subjects$ \COMMENT{Update character registry}
        \STATE SaveCheckpoint($image\_prompts$, $checkpoint\_file$)
    \ENDIF
\ENDFOR
\STATE SaveToDatabase($game\_uuid$, $day$, $image\_prompts$, $subjects$, $theme$)
\STATE $audio\_texts \gets$ Extract dialog texts
\STATE $job\_id \gets$ "\{game\_name\}-\{day\}"
\STATE GenerateAudioBatch($audio\_texts$, $job\_id$) \COMMENT{Azure TTS}
\end{algorithmic}
\end{mdframed}

\end{document}